\documentclass[11pt,a4paper]{article}
\usepackage[hyperref]{emnlp2020}
\usepackage{times}
\usepackage{latexsym}
\usepackage{multirow}
\usepackage{booktabs}
\usepackage{graphicx}
\usepackage{subcaption}

\usepackage{microtype}

\aclfinalcopy %
\def\aclpaperid{1792} %

\definecolor{mygreen}{rgb}{0, 0.6, 0}

\title{``I'd rather just go to bed'': Understanding Indirect Answers}

\author{Annie Louis\\
  Google Research, UK  \\
  {\tt annielouis@google.com}\\
  \And
  Dan Roth\thanks{* Work done at Google}\\
  University of Pennsylvania \\
  {\tt danroth@seas.upenn.edu}\\
\AND
  Filip Radlinski \\
  Google Research, UK\\
  {\tt filiprad@google.com}\\
  }

\date{}

\begin{document}
\maketitle
\begin{abstract}

We revisit a pragmatic inference problem in dialog: Understanding  indirect responses to questions. Humans can interpret \emph{`I'm starving.'} in response to \emph{`Hungry?'}, even without direct cue words such as `yes' and `no'. In dialog
systems, allowing natural responses rather than closed vocabularies would be similarly beneficial. However, today's systems
are only as sensitive to these pragmatic moves as their language model
allows. We create and 
release\footnote{The corpus can be downloaded from \url{http://goo.gle/circa}} the first large-scale English language corpus `Circa' with 34,268
(polar question, indirect answer) pairs to enable progress on this task. 
The data was collected via elaborate crowd-sourcing, 
and contains utterances with 
yes/no meaning, as well as uncertain, middle-ground, and 
conditional responses. We also present BERT-based neural models  to 
predict such categories for a question-answer pair. We find that while
transfer learning from entailment works reasonably, performance is not yet sufficient for robust dialog. 
Our models reach  82-88\% accuracy for a 4-class distinction, and 74-85\% for 6 classes.
\end{abstract}

\section{Introduction}

Humans produce and interpret complex utterances even in simple 
scenarios. For example, for the polar (yes/no) question 
\emph{`Want to get dinner?'}, there are many 
perfectly natural responses in addition to `yes' and `no', as in Table \ref{tab:example}. How should
a dialog system interpret these {\sc indirect} answers? Many
can be understood based on the answer text alone, e.g. \emph{`I would like 
that'} or \emph{`I'd rather just go to bed'}. For others, the
question is crucial, e.g.~\emph{`Dinner would be lovely.'} is a positive
reply here, but a negative answer to \emph{`Want to get lunch?'}. 
In this paper, we present the first large scale corpus and models for
interpreting such indirect answers. 

Previous attempts to interpret 
indirect yes/no answers have been small scale and without 
data-driven techniques \cite{green1999interpreting,de2010good}. 
However, recent success on many language understanding problems \cite{wang1804glue},
the impressive generation capabilities of modern dialog systems \cite{zhang2019dialogpt,adiwardana2020towards},
as well as the huge interest in yes/no question-answering 
\cite{choi-etal-2018-quac,clark-etal-2019-boolq} have created a conducive
environment for revisiting this hard task.

\begin{table}[t]
\begin{footnotesize}
\begin{tabular}{|p{7.2cm}|}
\hline
\multicolumn{1}{|c|}{\bf ``Want to get some dinner together?''} \\ 
``I know a restaurant we could get a reservation at.'' \\
``I have already eaten recently.''\\
``I hope to make it home by supper but I'm not sure I can.''\\
``Dinner would be lovely.''\\
``I'd rather just go to bed.''\\
``There's a few new restaurants we could go to.''\\
``I would like that.''\\
``We could do dinner this weekend.''\\
``I would like to go somewhere casual.''\\
``I'd like to try the new Italian place.''\\
\hline
\end{tabular}
\end{footnotesize}
\caption{A polar question with 10 indirect responses, taken from our corpus.}
\label{tab:example}
\end{table}

We introduce \emph{Circa}, a new dataset with 34K pairs of polar questions and 
indirect answers in the English language. This high quality corpus consists of natural responses collected
via crowd workers, and goes beyond binary
yes/no meaning to include conditionals, uncertain, and middle-ground answers.
Circa contains many phenomena of interest, although the first step, which we address here, is how to robustly classify a question-answer pair into one of the above meaning categories. 
We find that BERT \cite{devlin-etal-2019-bert} fine-tuned on entailment data is an effective
initial approach,
mirroring the success in question-answering work involving yes/no questions \cite{clark-etal-2019-boolq}. 
It reaches an accuracy of 85-88\% for responses 
in the same situational context, and 6-10\% lower accuracy on held-out scenarios.  The answer 
text itself (as in \emph{`I would like that'}) contains strong cues leading to 78-82\% accuracy, however the best results come from
jointly analyzing the question and the answer.

\section{Related work}
\label{sec:related_work}

Indirect answers to polar questions 
are reasonably frequent and warrant deep study.
Early work put the proportion at 13\% in 
face-to-face and telephone conversations \cite{stenstrom1984questions}, and at 27\% in an 
instruction giving/following map task \cite{rossen1997yes,hockey1997can}. For a more
recent and larger analysis, consider the Cornell Movie Dialog Corpus \cite{cornellmovie}. We heuristically mined yes/no questions and their following utterances, finding 6,327 pairs. 
Direct answers (i.e., answers with `yes', `no', `maybe' and related terms such as `okay', `yup', etc.) only 
cover 53\% of the pairs. This suggests that indirect responses could 
be even more frequent in natural open-domain dialogue.

Even when a direct answer is possible, speakers use
indirect answers to be cooperative and address
anticipated follow-up questions \cite{stenstrom1984questions}, to provide explanations in the case of a negative answer \cite{stenstrom1984questions}, to 
block misleading interpretations that may arise from a
curt `yes' or `no' reply \cite{hirschberg}, 
and since indirect answers may appear more polite 
\cite{brown1978universals}. But we lack a large corpus of
such answers to study these multiple pragmatic functions. Our work aims to
fill this gap.

On the computational side, there are impressive efforts towards planning, generation, 
and detection of indirect answers, albeit, on a small scale, and without 
data-driven approaches. \newcite{green1999interpreting}'s
early work leverages discourse relations for generating indirect answers.
For example, an `elaboration' may be relevant for a `yes'
response, and a `contrast' might help convey a `no' answer. \newcite{de2009not} 
reason about such answers using Markov Logic Networks. In subsequent work, \newcite{de2010good}
present one of the first data-driven studies into indirect answers containing scalar
adjectives. They mine a set of 224 question-answer pairs from interview transcripts and
dialog corpora.
Using polarity information from review data, and manual 
coding of test samples, they achieve an accuracy of 71\% on three classes `yes', `no' and `uncertain'. 
Our work aims to collect 
a much larger and more diverse natural corpus, and 
demonstrates the first automatic approach using recent advances in 
natural language inference (NLI). \newcite{de2010good} also
demonstrated the first crowd annotation of indirect answers, and 
we draw on many aspects of their formulation for the
creation of our corpus. 

An unexpected limelight on yes/no questions has also arisen
in recent question-answering (QA) work. Researchers
have noticed that yes/no questions are complex and naturally arise (as high as 20\%) when questions 
are posed one after the other in a conversation 
\cite{reddy2019coqa,choi-etal-2018-quac}. 
Their goal is to produce direct
`yes' or `no' answers, but obtaining them requires inference
against a paragraph or excerpt, an analogous task to our yes/no inference
from indirect answers. Very 
recent work \cite{clark-etal-2019-boolq} has specifically
sought to improve this ability in QA systems, by building a corpus
of 16K yes/no factual questions paired with Wikipedia passages from
which the yes/no answer can be inferred. Departing from factual 
texts, our focus is on single-turn \emph{conversational} responses in everyday situations. 
The latter are faithful, cooperative, and grounded in 
world knowledge. Still, transfer learning from factual corpora 
could prove useful and we explore this too.

\section{The Circa Corpus}

\begin{table}[ht]
\footnotesize
\begin{tabular}{|p{7cm}|}
\hline
{\bf S1: Talking to a friend about food preferences.} \newline
Q: ``Do you like pizza?'' \newline     
A: ``I like it when the toppings are meat, not vegetable.'' \\\hline
{\bf S2: Talking to a friend about music preferences.} \newline
Q: ``Do you like guitars?''  \newline
A: ``I practice playing each weekend.'' \\\hline
{\bf S3: Talking to a friend about weekend activities.} \newline
Q: ``Are you available this Sunday evening?'' \newline
A: ``What did you have in mind?'' \\ \hline
{\bf S4: Talking to a friend about book preferences.} \newline
Q: ``Are you a fan of comic books?''\newline
A: ``I read an Archie every time I have lunch.'' \\\hline
{\bf S5. Your friend is visiting from out of town.} \newline
Q: ``Would you like to go out for dinner?''  \newline
A: ``I could go for some Mexican.'' \\\hline
{\bf S6. Two colleagues leaving work on a Friday.} \newline
Q: ``Long week?'' \newline
A: ``I've had worse weeks.'' \\ \hline
{\bf S7. You friend is planning to buy a flat in New York.} \newline
Q: ``Does the flat's price fit your long-term budget?''    \newline
A: ``Well, if it doesn't I will definitely refinance my mortgage.'' \\\hline
{\bf S8. Your friend is thinking of switching jobs.}\newline
Q: ``Do you have to travel far?''\newline
A: ``My commute is about 10 minutes.'' \\\hline
{\bf S9. Two childhood neighbours uexpectedly run into each other at a cafe.} \newline
Q: ``Are you going to the high school reunion in June?'' \newline
A: ``I forgot all about that.''\\ \hline
{\bf S10. Meeting your new neighbour for the first time.} \newline
Q: ``Did you move from near-by?'' \newline
A: ``I am from Canada.'' \\\hline
\end{tabular}
\caption{Examples of questions and answers in our 10 dialogue scenarios.}
\label{tab:qaexamples}
\end{table}

Circa (meaning approximately) is our crowd-sourced corpus for research on indirect answers.
It contains 34,268 question-answer pairs, comprising
3,431 unique questions with up to 10 answers to each. Each 
question-answer pair is also annotated with meaning based on a fixed
set of categories.\footnote{The data can be downloaded from \url{http://goo.gle/circa}} We explain 
all aspects of the collection process below, but in the interest of space, 
further details (complete annotator instructions, interfaces, and examples)
are in the Appendix.

Our data consists of very short dialogues, each containing a question and its indirect answer. The 
texts are varied in semantic and syntactic forms, and grounded in 10 different
situational contexts. Table \ref{tab:qaexamples} presents some examples, showing that the answers go beyond binary yes/no distinctions. 
We created this data via an elaborate crowd-annotation exercise which comprised of \emph{4 steps} described next. 

\subsection{Step 1: Dialog scenarios} We designed 10 diverse prompts to serve as situational contexts. 
These everyday situations can lead to productive dialog, and 
simple yes/no questions (i.e., they do not depend on elaborate prior 
context for initiating a conversation). We designed them manually to intentionally cover a number of situations that encourage variety in questions and responses. As only a small number were needed, and the desiderata are not trivial, a crowd task was not suitable for prompt development. 
In Table \ref{tab:qaexamples}, S1--S10 are the titles of the 10 dialog scenarios (each also consisting of a longer description). 

The rest of the data were collected using crowd workers. 
We ran pilots for each step of data collection, and perused their results manually
to ensure clarity in guidelines, and quality of the data. 
We recruited native English speakers, mostly
from the USA, and a few from the UK and Canada. We did not collect any further information about the crowd workers.

\subsection{Step 2: Question collection} 
In this phase, we ask annotators to write 
yes/no questions for a given dialog scenario. 

\begin{table}
\begin{footnotesize}
\begin{tabular}{p{7cm}}
{\bf Annotator instructions} \\
In this task, we will ask you to provide yes/no questions in a social dialogue situation. \\
{\bf Example:} Suppose that you are trying to learn about a friend's movie preferences, 
but can only ask yes/no questions. Provide 5 useful questions that can be answered ``Yes'' or ``No''. \\
{\bf Note:} (1) We are looking for variety in the questions. For instance:\\
`Have you watched Star Wars?'\\
`Do you like movies with a complicated plot?' \\
`Did you enjoy the last movie we saw together?' \\
`Want to go watch a thriller this weekend?' \\
`Are you into the Avengers series?'\\
Note that the questions have different forms as well as different content. \\
(2) Remember that the setting is a conversation with a friend (or neighbour or colleague). 
Please keep the questions casual, so they would be natural during an informal conversation. \\
\end{tabular}
\end{footnotesize}
\caption{Annotator instructions for Step 2.}
\label{tab:question_collection}
\end{table}

Table \ref{tab:question_collection}
shows the instructions displayed to annotators.\footnote{In annotator instructions, we always use 
examples different from the actual scenarios in the exercise.} (Also see Figure \ref{fig:step1} and Table \ref{tab:scenarios_step1}.)
100 annotators each provided 5 questions
per scenario, resulting in 5,000 questions. 

Questions where annotators did not adhere to topic provided were removed. 
Of the remaining 4,710 questions, 84\% were unique. Understandably, some scenarios had more repetitions than others: 
For food preferences, 76\% of the questions were unique, as 
opposed to 95\% when talking about a friend's job change. Below we
show the most and least common questions
in the food context. 

\vspace{2mm}
\begin{footnotesize}
  \begin{tabular}{lp{5cm}}
    \multicolumn{2}{l}{\bf Most common food questions} \\
    \midrule
    25 times & Do you like spicy food? \\
    11 times & Are you vegetarian? \\
    10 times & Do you eat meat? \vspace{1mm} \\
    \multicolumn{2}{l}{\bf Sample of least common food questions}\\
    \midrule
    1 time & Have you ever tried vegan cuisine?\\
    1 time & Do you have a gluten allergy?\\
    1 time & Are you familiar with Thai food?\\
    \end{tabular}
    \end{footnotesize}
\vspace{1mm}

   The most common one \emph{`Do you like spicy food?'} was
suggested by 25 out of 100 annotators.

One important aspect of our design is that each annotator was asked to provide five questions at the same time. 
Obvious questions often showed up as the \emph{first} question, with later questions becoming more complex and more interesting. Question properties such as length, 
inverse document frequency, and type-token ratio confirm this difference.

\subsection{Step 3: Answer elicitation}
\label{sec:answer_elicitation}

We sampled 350 non-redundant questions per scenario, with
an equal number from the 5 question positions (see previous section)\footnote{For some scenarios, we obtained slightly fewer questions due to high repetitiveness at earlier question positions.}. %
Using a different set of annotators than Step 2, we then elicited 10 indirect answers for each question. The annotator instructions are provided in Table \ref{tab:answer_elicitation}. 
For faster annotation, and to encourage diverse answers, we displayed five questions (in same situational context) simultaneously. 
The five questions were chosen to be diverse (based on 
cosine similarity between nouns and adjectives in the questions), and annotators were instructed to 
treat them independently.\footnote{Note that the 10 answers for each question in the Circa corpus are always from 10 different annotators.}
See Figure \ref{fig:step2} for an example display.

Importantly, a key design consideration was that we do \emph{not} instruct
annotators to produce an answer with a specific meaning. Rather 
the annotator composes a natural
response without reflecting upon a required meaning. 
We believe 
this flexibility is important to ensure that varied meanings (even ambiguous ones)
are present in our data. This is verified in our analysis in Section \ref{sec:label_distributions}. 

Table \ref{tab:example} provides example answers from this
step.  Note that the 10 answers have varied meanings departing
from definite `yes' and `no'. The answers were high quality and diverse in form. 83\% of
answers appear only once in the corpus. 
At the same time, there are 
a few prototypical answers for the different meanings. 
Below we show the most repeated responses (and frequency)
for `yes' and `no' meanings, and also for cases where the answer is conditional 
upon some situation, and middle ground responses. 

\vspace{1mm}
\begin{footnotesize}
    \begin{tabular}{rp{2.7cm} | rp{2.3cm}}
    \multicolumn{2}{l|}{\bf `yes' answers}    & \multicolumn{2}{l}{\bf `no' answers }\\    
     \midrule
     59 & I would love to          & 21 & I don't drink \\
     40 & Let's do it             & 18 & I prefer pop\\
     40 & That would be good       & 18 & I wish! \\
\\
    \multicolumn{2}{l|}{\bf `conditional yes'}  & \multicolumn{2}{l}{\bf `middle-ground' }\\
    \midrule
    9 & If the weather is nice    &  14 & I'm not sure yet\\
    6 & If I can afford it & 10 & Which one?\\
    6 & Depends where you want to go          & 9 & It's OK\\
    \end{tabular}
    \end{footnotesize}
    
    These responses
indicate that strong lexical signals for meaning are present in the answer.

\begin{table}
\begin{footnotesize}
\begin{tabular}{p{7.2cm}}
{\bf Annotator Instructions}\\
You will be given a social situation, for example, talking to your friend or neighbour. \\
{\bf Task:} You will be asked to respond to a question from your friend/neighbour but without using the words ‘yes’ or ‘no’ (or similar words like ‘yeah’, etc). Please provide a possible answer, it does not have to be your real opinion. Rather you should provide a possible answer from which your friend will be able to infer whether you mean ‘yes’, ‘no’, ‘maybe’ or ‘sometimes’. \\
{\bf Example}: Here are three such answers to your friend’s question about movies.\\ 
\emph{Do you like movies with sad endings?} \\
(a) I often watch them. (Meaning=Yes) \\
(b) I prefer movies which make me laugh. (Meaning=No)\\
(c) When the plot is also good. (Meaning=Sometimes) \\
\end{tabular}
\end{footnotesize}
\caption{Annotator instructions for Step 3.}
\label{tab:answer_elicitation}
\end{table}

\subsection{Step 4: Marking interpretations}
\label{sec:interpretation}

\begin{table}[t]
\begin{footnotesize}
    \begin{tabular}{p{7.2cm}}
{\bf Annotator instructions}\\
 You will be shown short dialogues between two friends/colleagues (X and Y) in a certain context. For example: \\
Context: X wants to know about Y's movie preferences. \\
X: "Do you like movies with sad endings?" \\
Y: ``I often watch them." \\
In all the dialogues, X asks a simple `Yes/No' question, and Y answers it with a short sentence or a phrase. \\
{\bf Task:} We need your help to interpret Y's answer. Read the dialog and tell us how you think X will interpret Y's answer. Your options are: 
\emph{X will think that Y means} \\
(1) `yes' \\
(2) `probably yes' / `sometimes yes'\\
(3) `yes, subject to some conditions'\\
(4) `no'\\
(5) `probably no'\\
(6) `in the middle, neither yes nor no'\\
(7) `I am not sure how X will interpret Y's answer' \\
If Y's response does not fit any of the above, please choose the option (8) `Other', and leave a short comment. \\
For our example above, the likely interpretation is `yes'. 
    \end{tabular}
\end{footnotesize}
    \caption{Annotator instructions for Step 4.}
    \label{tab:interpretation_annotation}
\end{table}

\begin{table*}[h]
\begin{footnotesize}
\begin{tabular}{|p{4.8cm}| p{4.8cm}| p{4.8cm}|}
\hline
{\bf Yes} \newline
Q:  Do you have any pets?\newline     
A: My cat just turned one year old.
& 
{\bf Probably yes / sometimes yes} \newline
Q: Do you like mysteries?  \newline
A: I have a few that I like.
& 
{\bf Yes, subject to some conditions} \newline
Q:  Do you enjoy drum solos?\newline
A:  When someone's a master.\\ \hline
{\bf No} \newline
Q: Do you have a house?\newline
A: We are in a 9th floor apartment. 
&
{\bf Probably no} \newline
Q: Are you interested in fishing this weekend? A: It's supposed to rain.
&
{\bf In the middle} \newline
Q: Did you find this week good?\newline
A: It was the same as always.\\ \hline
\end{tabular}
\end{footnotesize}
\caption{Example question and answer pairs where all 5 annotators agreed
on the label.}
\label{tab:high_agreement_examples}
\end{table*}

Finally, we ask a third set of annotators to mark interpretations for all the QA (question, indirect answer)  pairs 
from Step 2. 
In particular, they are asked to judge how the question-seeker would interpret the
answer provided. As in most NLP tasks, interpretations will vary, and so
we obtain \emph{five} annotations per pair. 

The correct label categories are not readily clear, but the variety of examples in our corpus made
it certain that just `yes' and `no' will not suffice. 
Building on prior work by \newcite{de2010good}, and a pilot experiment, we identified categories that can be annotated reliably. These
are shown in Table \ref{tab:interpretation_annotation}.
The annotators were asked to assume the dialogs were co-operative and casual. 
They were advised to use `probably yes/no' when they cannot infer a definite meaning. If
X will not infer a `yes' without some condition being met, then the class `yes, subject to 
some conditions' was to be chosen. 
Detailed instructions with exact phrasing,
and practice questions are in the appendix (Figure \ref{fig:step3}, Tables~\ref{tab:scenarios_step3} and \ref{tab:practice_step3}).
Annotators took an average of 23 seconds per question-answer pair. 

Annotators were also given an option to flag improper content. We remove those QA pairs which were flagged by even one
of the five annotators. The authors also read every question, and  used a blacklist of words for additional filtering.
The remaining 34,268 pairs comprise the final corpus.

\subsection{Gold standard labels}

Each (question, indirect answer) pair was marked by five annotators, so we use majority judgement as the gold standard, subject to \emph{at least three}
annotators making that judgement.

We use two  aggregation schemes. The {\sc strict} scheme keeps all eight class distinctions
from  Table \ref{tab:interpretation_annotation}. 
A more {\sc relaxed} label is computed by
collapsing the 
uncertain classes with the definite ones: `probably yes / sometimes yes' $\rightarrow$ `yes', 
`probably no' $\rightarrow$ `no', and `I am not sure how X will interpret Y's answer' $\rightarrow$
`In the middle, neither yes nor no'. 
These classes were commonly confused by the raters.  
The `Other' class was used mostly when the 
question was not polar (e.g, disjunctive ones such as \emph{`Do you like to dine-in or take-out?'}).
To illustrate the richness of the Circa corpus, 
Table \ref{tab:high_agreement_examples} shows one QA pair from 
each class. %

\subsection{Label distributions}
\label{sec:label_distributions}

We now analyze the 
distribution of the gold standard labels. For {\sc strict} labels (Table \ref{tab:strict_label_distributions}), 
only 8\% of the examples (marked `N/A')
do not receive a majority vote. The most frequent class is `yes' (42\% of the data). 
`No' is less frequent (32\%). 
The third most frequent is `conditional yes', indicating 
that conditional preferences may be common. 
The `probably' classes are around 3-4\%, each with over a thousand
examples in the corpus. There is also a notable number of `in the middle' examples. With the collapsed 
{\sc relaxed} labels (Table \ref{tab:relaxed_label_distributions}), the proportion of `yes' and `no' increase slightly, and `N/A' examples drop to only 2\%. 
These distributions reflect the rich patterns in 
our data.

\begin{table}
    \centering
    \begin{footnotesize}
    \begin{tabular}{p{4.2cm} | r r}
      \toprule
      {\bf Label}	& \multicolumn{2}{c}{\sc strict}\\
      \midrule
                                     Yes & 	14,504 & (42.3\%) \\ 
                                      No & 	10,829 & (31.6\%) \\ 
            Probably yes / sometimes yes & 	1,244  & (3.6\%) \\ 
         Yes, subject to some conditions & 	2,583  & (7.5\%)  \\ 
                             Probably no & 	1,160  & (3.4\%) \\ 
      In the middle, neither yes nor no  &    638  & (1.9\%)	 \\ 
                        I am not sure    &     63  & (0.2\%)  \\ 
                                  Other  &    504  & (1.5\%) \\ 
                                      N/A & 2,743  & (8.0\%) \\
                                      \bottomrule
    \end{tabular}
     \end{footnotesize}
    \caption{Distribution of {\sc strict} gold standard labels. 'N/A' indicates lack of majority agreement. }
    \label{tab:strict_label_distributions}
\end{table}

\begin{table}
    \centering
    \begin{footnotesize}
    \begin{tabular}{p{4cm} | r r}
    \toprule
      {\bf Label}	& \multicolumn{2}{c}{\sc relaxed} \\
      \midrule
                                     Yes & 	16,628 & (48.5\%) \\ 
                                      No &	12,833 & (37.5\%) \\ 
         Yes, subject to some conditions & 	 2,583 & (7.5\%) \\ 
      In the middle, neither yes nor no	 &     949 & (2.8\%) \\ 
                                  Other & 	   504 & (1.5\%) \\ 
                                      N/A 	& 771 & (2.2\%) \\ 
                                      \bottomrule
    \end{tabular}
    \end{footnotesize}
    \caption{Distribution of {\sc relaxed} gold standard labels. 'N/A' indicates lack of majority agreement.}
    \label{tab:relaxed_label_distributions}
\end{table}
\subsection{Annotator agreement}

The Fleiss kappa scores are 0.61 for {\sc strict} and 0.76 for {\sc relaxed} labels (p-values $<$ 0.0001) indicating
substantial agreement. In fact, full agreement, as in Table \ref{tab:high_agreement_examples} where all 
five annotators agree on the {\sc strict} class, occurs for 49\% of pairs, a high proportion given the complexity of the 
task. When the labels are {\sc relaxed}, this
reaches  71\%. %
The full agreement distributions are:  

\begin{center}
    \begin{footnotesize}
    \begin{tabular}{l r  r}
\toprule
{\bf Agreement} & {\sc strict} & {\sc relaxed}\\ 
\midrule 
5 annotators	& 49.1\% & 71.0\% \\
4 annotators	& 23.8\% & 17.2\% \\
3 annotators	& 19.1\% & 9.6\% \\ 
\bottomrule
\end{tabular}
\end{footnotesize}
\end{center}

\subsection{Dialog scenarios and question type}
\label{sec:question_scenario_analysis}

As one would expect, different scenarios prompt
different types of questions and answers, and hence different 
label distributions. For the `friend switching jobs' scenario, 54.5\% of elicited
answers are have `yes' meaning ({\sc relaxed} labels). For book preferences, this is only 42\%.

Perhaps more unexpected is that a few questions have labels predominantly 
of the same kind. For example, all 10 answers to  \emph{`Ready for the weekend?'} receive a 
`yes' label, and they are all `no' for \emph{`Are you offering
the asking price?'}. While the first question is largely rhetorical, the second involves common sense: Most people negotiate real-estate prices. We found that 3\% of questions (95) have all answers with the same label, and for another 20\%, 8-9 answers have the same label. A model would still
need to identify the label as either `yes' or `no', but these skews indicate that there may
be some (weak) signals for the label in the question itself. After all, \emph{who isn't ready for the weekend?}

\subsection{Annotation Protocol}
While our annotation method is comprehensive, one might wonder how alternative approaches would have fared. 
We have not performed controlled tests of different approaches but we briefly document our choices to aid future work.

We performed pilot annotations for each step of our process. Our goal was to start with less restrictive settings, where annotators are given minimal and simple instructions. If we did not receive quality responses, we intended to give more specific directions. 
For example, in an initial pilot, sometimes annotators gave long answers which may be unrealistic in conversations eg. \emph{`I can tell you, without a shadow of a doubt, that there are very few things that I enjoy more than sitting in front of my computer.'} So in later annotations for all tasks, we added instructions `to keep the conversation casual’ which at least reminds annotators about this need. We performed spot checks on the results, but did not perform controlled tests. 

For answer elicitation, our simple instruction pilot produced answers with diverse meanings. The distribution of meanings also varied according to the question (some questions had a mix of meanings, others were skewed towards a few meanings).  So we decided against explicitly asking annotators to provide a certain meaning as that would create uniform meaning distributions, which may turn out unrealistic and miss dominant tendencies. We also considered that the explicit approach may reduce answer quality (for example, if a person who eats meat were to have to answer indirectly that they are vegetarian, they may be more likely to get their facts wrong, or try to rush through the task). 
 In fixing our choice, we took care to ensure that quality was not affected, and answers were varied. 

\section{Learning Task}

Obviously answers contain 
numerous cues for subsequent dialog flow, but in this first work we focus on 
meaning prediction. Specifically, given a question-answer pair, we 
classify it into one of the meaning categories in Tables 
\ref{tab:strict_label_distributions} and \ref{tab:relaxed_label_distributions}.

We consider two experimental settings: In the \emph{matched} setup, we assume that the response scenario is seen during training (randomly dividing our corpus examples into 60\% training, 20\% each for development/test). 
The \emph{unmatched} setting is aimed at understanding the performance on
unseen scenarios, i.e., whether models can learn the
semantics of yes/no answers beyond the patterns specific to individual scenarios. 
As our data contains 10 scenarios, we carry out 10 leave-one-out tasks, each time holding out one  
scenario (for example, `buying a flat in New York') as the test data, and use the remaining nine for 
training and development. %

For both the \emph{matched} and \emph{unmatched} setting, we consider two variants of the classification
problem: {\sc strict} (with 6 different labels, namely all except `other' and `N/A' in Table \ref{tab:strict_label_distributions}) and {\sc relaxed} with 4 labels (Table \ref{tab:relaxed_label_distributions}). We ignore %
the examples without a majority label and also those marked `unsure' or `other'. 
Thus our experiment data sizes are:

\vspace{1mm}
\begin{center}
\begin{small}
    \begin{tabular}{p{3.5cm} r r r}
    \toprule
    {\bf Experimental Setting}                    & \multicolumn{1}{c}{\bf Train} & \multicolumn{1}{c}{\bf Dev.} & \multicolumn{1}{c}{\bf Test} \\ 
    \midrule
    {\sc strict}-matched             & 18,574           & 6,192  & 6,192 \\
    {\sc relaxed}-matched            & 19,795           & 6,599  & 6,599  \\
    {\sc strict}-umatched       &  24,746  &  3,115  & 3,095\\
    {\sc relaxed}-unmatched     &  26,404  &  3,289   & 3,299 \\ 
    \bottomrule
    \end{tabular}
    \end{small}
\end{center}

For the \emph{unmatched} setting, these sizes are the average across the 10 leave-one-out sets.
\section{Models}
Building upon recent NLI systems, our approach leverages representations from 
unsupervised pre-training, and finetunes a multiclass classifier
over the BERT model \cite{devlin-etal-2019-bert}. However, we first consider other models for related tasks.

\subsection{Related baselines and corpora}
\label{sec:related_tasks}

\noindent {\bf BOOLQ} is a question-answering dataset focused on factual yes/no questions \cite{clark-etal-2019-boolq}. 
Here yes/no questions from web search queries are paired with Wikipedia paragraphs that help answer the question. 
There are  9.4k train, 3.2k development, and 3.2k test set examples, with two target classes, namely `yes' and `no'.
We train our own BOOLQ model with BERT pre-training. It reaches a development accuracy of 74.1\%\footnote{\newcite{clark-etal-2019-boolq} report
78\% (dev.) with BERT-Large.}.  This model only predicts two classes `yes' and `no'.

\vspace{1mm}
\noindent{\bf MNLI.} The MultiNLI corpus \cite{mnli} is a large corpus for textual
entailment. It consists of premise-hypothesis sentence pairs which are marked with three target
classes `entailment', `contradiction' and `neutral'. There are 392K train, 9K dev., and 
9K test examples.

Although not applicable to all indirect answers, semantic
consequence can be leveraged for interpreting some of them. 
For the question 
(Q) \emph{`Do you like Italian food?'}, consider two possible
answers (A) \emph{`I love Tuscan food.'} and (B) \emph{`I prefer
Mexican cuisine.'} . Let Q$'$ be the declarative (positive) sentence
derived from Q
i.e. Q$'$ = \emph{`I like Italian food'}. (Q$'$ can be obtained by inverting the subject
and auxiliary, and changing pronoun to first person). The meaning of A and B above, 
can then be obtained from an entailment system: A $\Longrightarrow$ Q$'$, hence `yes', while B contradicts Q$'$, hence `no'.

We thus obtain predictions from an MNLI system, and 
map the predicted three NLI classes in a post-processing step: `contradiction' $\rightarrow$ `no', 
`entailment' $\rightarrow$ `yes', and `neutral' $\rightarrow$ `in the middle'. Note that this approach cannot 
predict all the classes in the corpus. Before prediction, we rewrite our questions into
declarative form using syntactic rules on a constituency parse. Performance is much worse without this rewriting. 
Our models for the MNLI task start from a BERT checkpoint and 
reach a development accuracy of 84\%. 

This model improves with the syntactic rewriting of questions which we are able to do fairly accurately. 
We based the rewrite 
rules on 50 questions. On a different set of 50 questions, 38  
were rewritten fully accurately (manual inspection). Some errors arose from incorrect 
parsing and some are due to deficient rules. For example, we do not handle verb re-inflection. So \emph{`Did you enjoy 
the movie?'} gets rewritten into \emph{`I did enjoy the movie.’} rather than \emph{`I enjoyed the movie.'}  This rewriting
technique helped only the MNLI model. For other finetuning based models, which involve training, the models are able to learn from the question form itself.

\vspace{1mm}
\noindent{\bf Majority baseline.} This method predicts the most frequent label, `yes', for all examples. 

\subsection{Training with Question or Answer only}

\paragraph*{\bf Answer only.} In many cases, the answer to a question suffices for predicting the label
(see Table \ref{tab:example} and Section \ref{sec:answer_elicitation}); for example \emph{``I would like that.''} or
\emph{``I wish!''}.
This {\em Answer only} model  fine-tunes BERT to predict the class
based only on the answer. Similar experiments are done on NLI datasets to test if the hypothesis alone is at times sufficient for entailment prediction \cite{poliak-etal-2018-hypothesis,gururangan-etal-2018-annotation}. 
Such results are problematic for entailment (since it is \emph{defined} to depend on the truth of the premise). In contrast, our problem is primarily about the meaning of answers. This experiment will provide insight into the cues within indirect responses, an aspect 
not understood so far.

\vspace{1mm}
\noindent{\bf Question only.} Some questions commonly elicit certain answers (see Section \ref{sec:question_scenario_analysis}). These models test 
how well the question predicts the label. 

\subsection{Question-Answer Pair models}

These models take both the question and the answer. They all finetune BERT checkpoints, and the
question-answer pair is passed with a separator. 

\vspace{1mm}
\noindent{\bf BERT-YN} is BERT finetuned only on our Circa corpus (YN). 

We also explore how to transfer the strength of parameters learned for three related inference tasks.  

\vspace{1mm}
\noindent{\bf BERT-BOOLQ-YN} finetunes a BOOLQ model checkpoint (see Section \ref{sec:related_tasks}) on our corpus, 
with a new output layer. Since BOOLQ is a Yes/No question answering system, even if developed
for a different domain, we
expect to learn many semantics of yes/no answers from this data. 

\vspace{1mm}
\noindent{\bf BERT-MNLI-YN} is first fine-tuned on the MNLI corpus, followed by our YN data. This configuration tests if
the signals we hoped to capture with the out-of-the-box MNLI model (Section \ref{sec:related_tasks})
can be strengthened by training on our target task.

\vspace{1mm}
\noindent{\bf BERT-DIS-YN.} As discussed, indirect answers
also have discourse relations with the speaker's intent \cite{green1999interpreting}.
We implement this idea via a discourse connective prediction task. 
Consider again the texts from Section \ref{sec:related_tasks}: The likely connective between 
Q$'$ and A is `because' as in \emph{`I like Italian food [because] I love Tuscan food.'}. For Q$'$ and B, 
`but' would be more reasonable: \emph{`I like Italian food [but] I prefer Mexican cuisine.'}. We hypothesize that these
discourse relations will help learn the yes/no meaning via transfer learning. 

We use 400K examples (matching the MNLI data size) of explicit connectives and their arguments (a subset of \newcite{dissent}). 
We aim to predict the 5 connectives (because, but, if, when, and) based on their arguments. This task itself can be done 
with a development accuracy of 82\%. The best checkpoint is then finetuned on YN data.

\begin{table*}[h!]
    \centering
\begin{small}
    \begin{tabular}{l|r r|r r r r| r r r r}
\toprule
                 &  \multicolumn{6}{c|}{\bf Matched setting} & \multicolumn{4}{c}{\bf Unmatched setting} \\
    {\bf Model} & \multicolumn{2}{c|}{\bf Accuracy} & \multicolumn{4}{c|}{\bf Test F-Score} &  \multicolumn{4}{c}{\bf Test Accuracy} \\  
        &  {\bf Dev.} & {\bf Test} & {\bf Yes} & {\bf No} & {\bf C.Yes} & {\bf Mid} & {\bf Mean} & {\bf Std.} & {\bf Min.} & {\bf Max.} \\ \hline
         \multicolumn{11}{c}{}\\ 
          \multicolumn{11}{c}{\bf Baselines (no finetuning)}\\ \hline
    Majority class           & 50.2 & 49.3  &     66.0   &  0.0       & 0.0        & 0.0        & 50.4      & 4.3     & 43.6  & 56.8\\      
    MNLI                     & 28.4 & 28.9  &  34.4       & 52.8      & 0.0        & 6.9        & 28.1      & 2.8     & 24.2  & 34.1\\
    BOOLQ                    & 64.2 & {\bf 62.7} & {\bf 71.1}  & {\bf 59.6}& 0.0        & 0.0        & {\bf 63.3}  & 2.7     & {\bf 58.3}  & {\bf 66.5}\vspace{2mm}\\
    
                        \multicolumn{11}{c}{\bf BERT finetuned on Question or on Answer}\\ \hline
    BERT-YN (Question only)  & 56.4& 56.0       & 63.1       & 54.1       & 9.1        & 1.0        & 53.3    & 2.9    & 48.0     & 58.4  \\
    BERT-YN (Answer only)    & 83.0& {\bf 81.7} & {\bf 83.9} & {\bf 80.3} & {\bf 88.9} & {\bf 18.6} & {\bf 80.1} & 5.8    & {\bf 71.4}   & {\bf 87.8}\vspace{2mm} \\ 
                        \multicolumn{11}{c}{\bf BERT finetuned on Question $+$ Answer}\\ \hline
    BERT-YN                  & 88.4& \underline{87.8}       & 89.8       & 87.9       & 89.9     & 28.2       & 85.5       & 3.9    & 79.0         & 90.2  \\
    BERT-MNLI-YN             & 89.6& {\bf 88.2} & {\bf 90.4} & {\bf 88.5} & 89.3       & 29.4       & {\bf 87.1} & 3.0    & {\bf 81.9}   & {\bf 90.3} \\ 
    BERT-DIS-YN              & 88.0& 87.4       & 89.4       & 87.4       & {\bf 90.0} & {\bf 35.2} & 85.5     & 3.5    & 78.9    & 89.4  \\ 
    BERT-BOOLQ-YN            & 87.7& 87.1       & 89.0       & 86.9       & 89.6       & 30.9       & 85.3     & 3.7    & 78.8    & 89.4  \\ \bottomrule
      \end{tabular}
    \end{small}
    \caption{Performance on the {\bf relaxed}  labels.  The highest value in each column is in bold. For matched
    setting, we show dev. and test accuracies, and also F-scores for the 4 labels (`yes', `no', `conditional yes' and `in the middle'). In unmatched
    setting, we report summaries of 10 leave-one-out experiments. \emph{BERT-YN is significantly better than `Answer only' (McNemar's test, p-value $<$ 1e-6); BERT-MNLI-YN is \underline{not} significantly better than BERT-YN }.
    }
    \label{tab:relaxedresults}
\end{table*}

\begin{table*}
    \centering
\begin{small}
    \begin{tabular}{l|r  r|r r r r r r| r r r r}
        \toprule
                &  \multicolumn{8}{c|}{\bf Matched setting} & \multicolumn{4}{c}{\bf Unmatched setting}\\
    {\bf Model} & \multicolumn{2}{c|}{\bf Accuracy} & \multicolumn{6}{c|}{\bf Test F-Score} &   \multicolumn{4}{c}{\bf Test Accuracy}\\  
        & {\bf Dev.} & {\bf Test} & {\bf Yes} & {\bf P.Yes} & {\bf C.Yes} & {\bf No} & {\bf P.No} & {\bf Mid} & {\bf Mean} & {\bf Std.} & {\bf Min.} & {\bf Max.} \\ \hline
         \multicolumn{13}{c}{}\\ 
          \multicolumn{13}{c}{\bf Baselines (no finetuning)}\\ \hline
    Majority class          & 47.5 & 47.0         & 63.9       & 0.0        & 0.0       & 0.0       & 0.0       & 0.0       & 46.9       & 3.9  & 40.0  & 52.3\\      
    MNLI                    & 26.3 & 27.4         & 36.6       & 0.0        & 0.0       & 53.0      & 0.0       & 4.9       & 26.4       & 3.2  & 21.7  & 32.7\\
    BOOLQ                   &  59.4& {\bf 59.2}   & {\bf 70.4} & 0.0        & 0.0       & {\bf 57.0}& 0.0       & 0.0       & {\bf 58.9}   & 3.0  & {\bf 53.8}  & {\bf 63.7}\vspace{2mm}\\ 
             \multicolumn{13}{c}{\bf BERT finetuned on Question or Answer}\\ \hline
    BERT-YN (Question)      & 53.7 &  52.8       &  62.3      &  3.2      &  19.7       & 51.1      & 0.0      & 4.7      & 49.4       & 4.0  & 41.9  & 56.7  \\ 
    BERT-YN (Answer)        & 77.3 &  {\bf 77.8} & {\bf 82.5} & {\bf 49.5}& {\bf 90.2} & {\bf 77.3}& {\bf 16.2}& {\bf 26.9} & {\bf 75.8} & 5.8  & {\bf 65.4}  & {\bf 82.8}\vspace{2mm}\\

                        \multicolumn{13}{c}{\bf BERT finetuned on Question $+$ Answer}\\ \hline
    BERT-YN                 & 83.6 &  \underline{84.0}       &  88.7      & 49.9      & {\bf 90.2}       & 85.4      & 18.6      & 42.6      & 81.2    & 4.6  & 71.8        & 85.6   \\ 
    BERT-MNLI-YN            & 85.0 &  \underline{\bf 84.8} &  {\bf 89.8}& {\bf 51.8}& 89.8 & {\bf 86.6}& 18.0      & 41.3      & {\bf 82.8}  & 4.0  & {\bf 74.4}  & {\bf 86.7} \\ 
    BERT-DIS-YN             & 83.8 &  83.3       &  87.9      & 50.2      & 90.5       & 84.1      & 21.2& {\bf 50.8} & 81.5   & 4.5  & 73.1        & 86.3  \\ 
    BERT-BOOLQ-YN           & 83.1 &  83.4       &  88.2      & 51.2      & 89.1       & 84.5      & {\bf 22.1}    & 43.7       & 81.1   & 4.3  & 73.3        & 85.8   \\ \bottomrule
    \end{tabular}
    \end{small}
    \caption{Performance on the {\bf strict}  labels.  The highest
    value in each column is in bold. \\\emph{BERT-YN is significantly better than `Answer only' (McNemar's test, p-value $<$ 1e-6), and BERT-MNLI-YN is better than BERT-YN (p-value $<$ 0.02).}}
    \label{tab:strictresults}
\end{table*}

\section{Experiments}

We use pre-trained BERT models (with 12 layers, 768 hidden units, and 12 attention heads, 110M parameters) 
for all our experiments. The experiments were done on a single Cloud TPU, and finetuning on our corpus 
takes under 30 minutes. 

\subsection{Setup and Hyperparameter Tuning}
For the base models (MNLI, BOOLQ, and DIS), we tuned the learning rate (values 5e-5, 3e-5, 2e-5), the number of epochs 
(2, 3, 4), and train batch size (16, 32) in
an exhaustive combination. For finetuning on yes/no data, we tune the learning rate
while setting the epochs to 3 and training batch size to 32. We also perform three random
restarts in each configuration. Performance was stable across the restarts (accuracy variation $\le 1\%$). So we 
take the best model on the development set as our final model. The best hyperparameters are in the Appendix.
For the unmatched setting, we do 10 leave-one-out tasks. Here we take the best parameters from the matched
setting, and use that configuration to train all 10 experiments.

\subsection{Results}
We report the development accuracy
and detailed test results in Table  \ref{tab:relaxedresults} ({\sc relaxed} setting) and Table~\ref{tab:strictresults} ({\sc strict} setting). For the unmatched setting, we report the mean accuracy and
standard deviation across the 10 folds, and the min and max values. 
We first discuss results for the {\sc relaxed} labels. The findings for {\sc strict} are similar. 

The majority baseline (`yes' class) leads to 
an accuracy of 49\%. The MNLI to yes/no label mapping (no finetuning) is reasonable in terms of F-score 
for the `no'
class, but is poor for `yes'. BOOLQ is the best baseline with 63\% accuracy. However,
there is no recall for labels other than `yes' and `no'. 

The question-only and answer-only results are noteworthy. 
The question-only model outperforms the majority baseline. 
On the other hand, the answer text contains
strong signals, with 82\%  accuracy, or about 20\% better than the best
baseline.

But models using both question and answer reach 5-6\% greater accuracy and significantly  outperform 
the answer-only model (McNemar's test, p-values $<$ 1e-6). As expected, these joint models are necessary when a string is a possible answer to multiple questions. An answer-only model is easily misled in these cases, as the examples below show:

\vspace{2mm}
\begin{footnotesize}
\begin{tabular}{p{6cm}}
(1)  ``Is there something you absolutely won't eat?''
``I really dislike bananas.''\\
Answer-only prediction: `No'\\
Question$+$Answer prediction: `Yes'\\
\\
(2) ``Do you need a nap?'' \\
``I have plenty of energy.'' \\
Answer-only prediction: `Yes'\\
Question$+$Answer prediction: `No''\\
\end{tabular}
\end{footnotesize}
\vspace{2mm}

The best F-scores are obtained by an MNLI transfer task, reaching 88.2\% accuracy in the matched setting.
But it is not signficantly better than a  no-transfer BERT-YN model (McNemar's test).

The unmatched setting shows that the models are worse 
when a scenario has not been seen in the training data. While it may not be possible for 
every conversational system to generalize across scenarios, a semantic classification such as yes/no
should ideally be robust to such changes. Instead we see a 6-10\% accuracy gap between 
the in-scenario test accuracies, and minimum out-of-scenario accuracy. The best accuracies are
reached by a MNLI model. The highest accuracy on a scenario is 90\% (`music preferences', `weekend activities'), 
and lowest is 82\% (`buying a flat in New York' and
`switching jobs'). The latter scenarios are quite different than the rest,
indicating scope for improving the models.

 The `in the middle' class has much worse results compared to the rest. The class has low frequency 
 but also comprises responses of different types. Uncertain responses such as \emph{`I am not sure.'}  
 appear easy to classify. But responses which do not take a stance: \emph{`Do you know if it's raining 
 outside? I'm prepared regardless.'} are harder. Sometimes, the interpretation 
 is left to the judgement of the listener. Eg. \emph{`travelling an hour away'} could be 
 interpreted as `far away’ or `close by’ depending on the context and perceptions of the listener. These cases need
 models to deeply connect the question and answer, and are missed by our technique.  

The general trends for {\sc strict} labels is similar: The best accuracy is again
reached with MNLI pretraining. It is 85\% for the matched case, a small but significant gain over
BERT-YN (McNemar's test, p-value $<$ 0.02). The accuracy is 10\% 
lower (74\%) for hardest experiment in leave-one-out. 

For {\sc strict} labels, the `probably no' class is hardest to predict even though it is as frequent as `probably yes' and 
close to double the size of the `in the middle' class. We found that 
`probably no' examples are heavily (69\%) mis-predicted into the `no' class. Utterances which explicitly convey
uncertainty eg. (\emph{`I don't believe so'}) or comparison (\emph{`Is everything good? Not the greatest.'}) are somewhat easier to predict. On the other hand, in: 

\vspace{2mm}
\begin{footnotesize}
\begin{tabular}{p{6cm}}
``Have you ever bought a romance novel?''
``I prefer to read horror books.''\\
Best model: ``No''\\
Gold standard: ``Probably no''\\
\end{tabular}
\end{footnotesize}
\vspace{2mm}

The speaker prefers horror genre, but it does not preclude ever buying a romance novel. These
subtleties are understandably harder for systems.

Overall, while MNLI based transfer learning has led to small improvements, incorporating the right
information for the task remains a challenge.

\section{Conclusion}

We have presented a new dataset containing 
natural indirect yes/no answers, as well as other significant pragmatic moves in the form of conditionals and uncertain utterances.
Our first approach towards automatic interpretation is promising, but there is a
significant gap especially for examples 
outside training scenarios. 
Our model does not yet classify additional information in responses (\emph{`Dinner? Let's go for Italian.'}
indicates not only a `yes' answer, but also a preference for Italian food.). Moreover, we have  explored the phenomena in 
English. There are exciting avenues for multilingual work to account for language and cultural differences.

\bibliographystyle{acl_natbib}
\bibliography{indirect}

\clearpage
\appendix

\title{Appendix}

\section{Hyperparameter settings}

For the base models (MNLI, BOOLQ, DIS), we tuned the learning rate (values 5e-5, 3e-5, 2e-5), the number of epochs
(2, 3, 4), and train batch size (16, 32) in
an exhaustive combination of these parameters. The best performance on development data was obtained with 
the following settings: 

\vspace{2mm}
\begin{small}
\begin{tabular}{l|c c c}
\toprule
    {\bf Model}  &  {\bf learning rate} & {\bf no. epochs} & {\bf batch size} \\ \hline
    MNLI    & 2e-5  & 3 & 16 \\
    BOOLQ   & 3e-5  & 4 & 16 \\
    DIS     & 2e-5  & 2 & 32 \\
    \bottomrule
      \end{tabular}
\end{small}
\vspace{2mm}

For finetuning on yes/no data (matched setting), we tune the learning rate
while setting the number of epochs to 3, and training batch size to 32. We also perform three random
restarts in each configuration. Performance was stable across the restarts (accuracy variation $\le 1\%$). We 
take the best model on the development set as our final model. The chosen learning rates are in the table
below:

\vspace{2mm}
\begin{small}
\begin{tabular}{l|c c}
\toprule
    {\bf Model}  &  {\sc strict} & {\sc relaxed} \\ \hline
    BERT-YN (Question only)  & 2e-5 & 3e-5\\
    BERT-YN (Answer only)    & 2e-5 & 2e-5\\
    BERT-YN                  & 3e-5 & 3e-5\\
    BERT-MNLI-YN             & 2e-5 & 5e-5\\ 
    BERT-DIS-YN              & 3e-5 & 5e-5\\
    BERT-BOOLQ-YN            & 3e-5 & 5e-5\\ 
    \bottomrule
      \end{tabular}
\end{small}
\vspace{2mm}

For the unmatched setting, we do 10 leave-one-out experiments. Here we use the best parameters from the matched
setting, and use the same configuration for training in all 10 experiments.

\section{Annotation Instructions}

We now detail the complete instructions to annotators, along with
interface examples, prompt texts, and practice items. 

\begin{figure*}[h!]
  \centering
  \begin{subfigure}[b]{0.45\linewidth}
    \fbox{\includegraphics[width=0.9\linewidth]{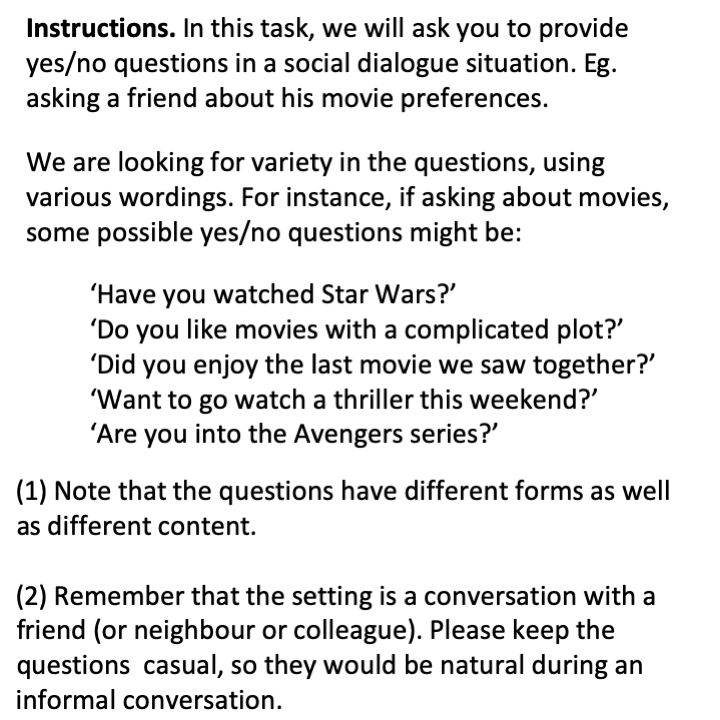}}
    \caption{Instructions}
  \end{subfigure}
  \begin{subfigure}[b]{0.45\linewidth}
    \fbox{\includegraphics[width=0.9\linewidth]{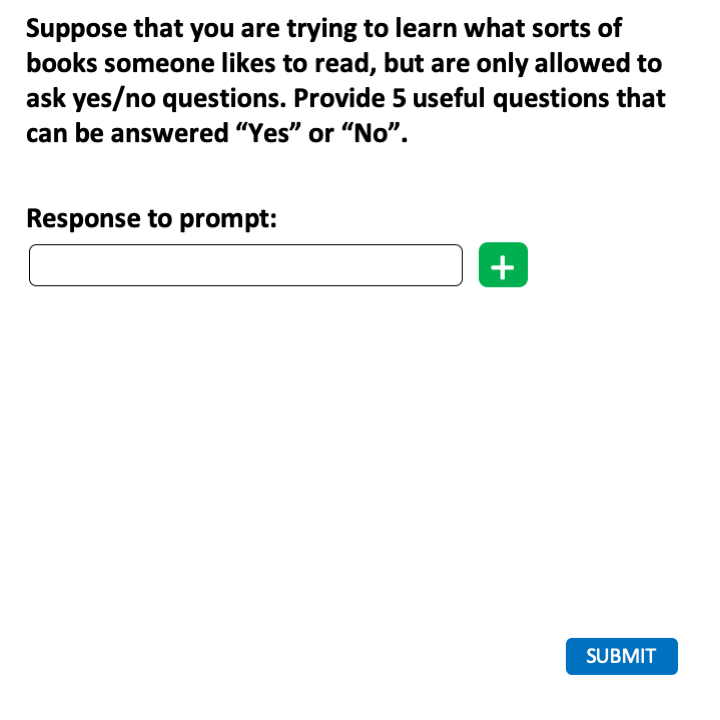}}
    \caption{Example item}
  \end{subfigure}
  \caption{Annotator interface for question collection.}
  \label{fig:step1}
\end{figure*}

\paragraph*{\bf Question collection.} In this step, an annotator is shown a scenario, and asked to provide five yes/no questions. The instructions, and interface for an example item are in Figure \ref{fig:step1}. The descriptions used for each of the 10 dialog scenarios are in Table \ref{tab:scenarios_step1}. 

\begin{figure*}
  \centering
  \begin{subfigure}[b]{0.45\linewidth}
    \fbox{\includegraphics[width=0.9\linewidth]{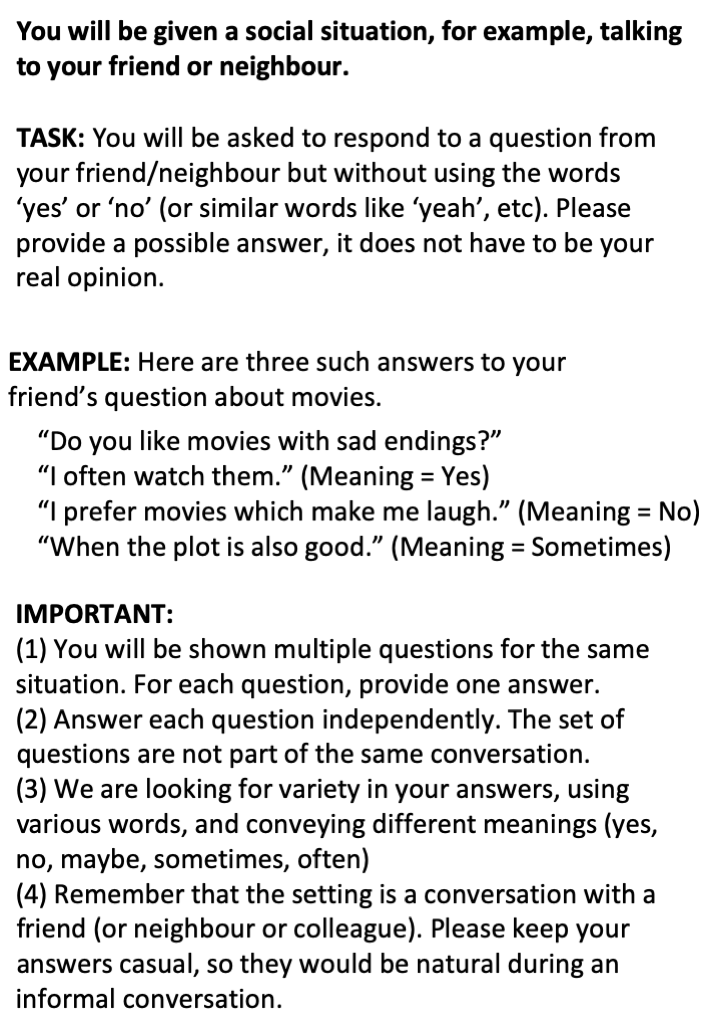}}
    \caption{Instructions}
  \end{subfigure}
  \begin{subfigure}[b]{0.45\linewidth}
    \fbox{\includegraphics[width=0.9\linewidth]{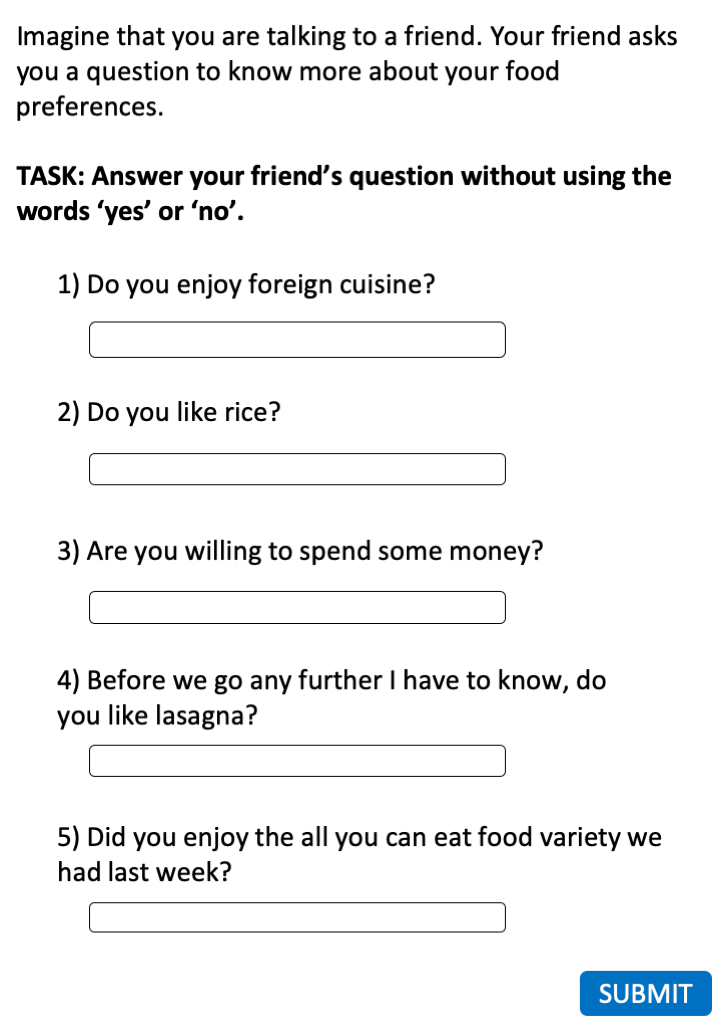}}
    \caption{Example item}
  \end{subfigure}
  \caption{Annotator interface for answer elicitation.}
  \label{fig:step2}
\end{figure*}

\paragraph*{\bf Answer elicitation.} Similarly, the instructions and interface for collecting answers are in Figure \ref{fig:step2}. Note that we show 5 questions on each screen, to reduce annotation time. The 5 questions are taken from the same scenario but such that they are not too similar. We enforce non-redundancy by keeping the pairwise similarity between any two questions on the same screen to less than 0.35 (measured by cosine similarity between the adjectives and nouns in the questions).

We have the same 10 scenarios but their descriptions are changed slightly to suit to the answer elicitation task. These prompts are in Table \ref{tab:scenarios_step2}.

\begin{figure*}
  \centering
  \begin{subfigure}[b]{0.45\linewidth}
    \fbox{\includegraphics[width=0.9\linewidth]{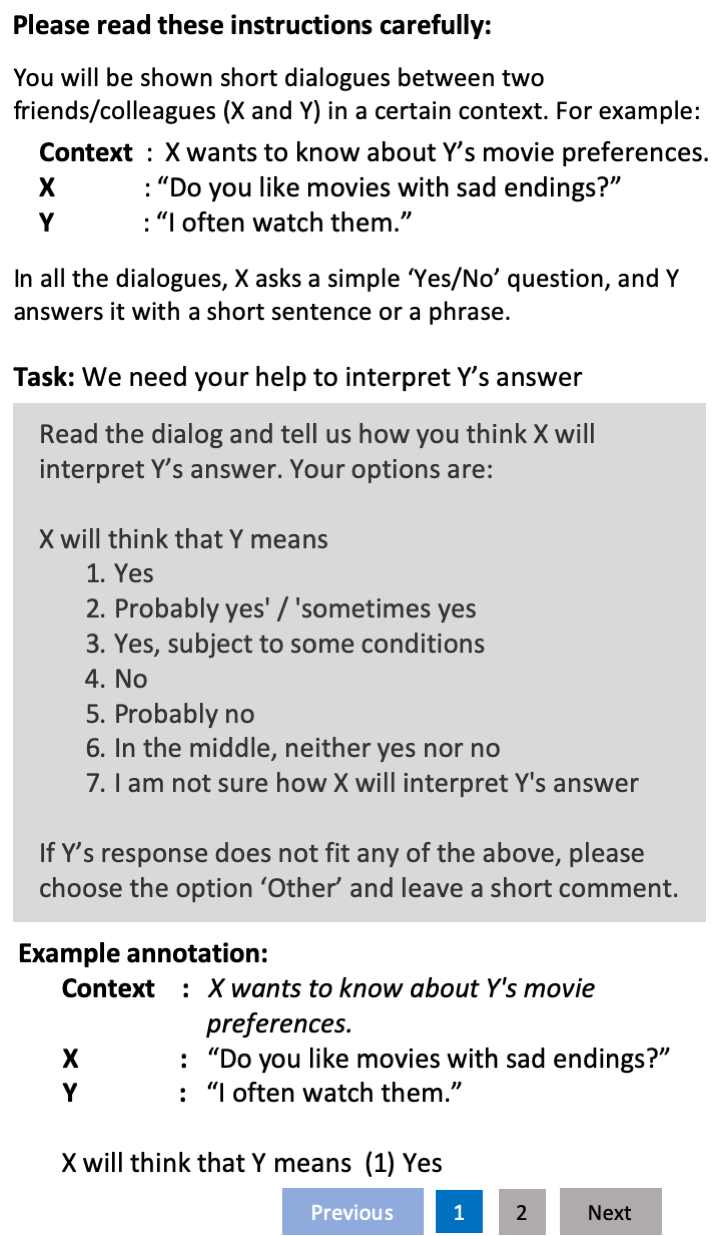}}
    \caption{Instructions}
  \end{subfigure}
  \begin{subfigure}[b]{0.45\linewidth}
    \fbox{\includegraphics[width=0.9\linewidth]{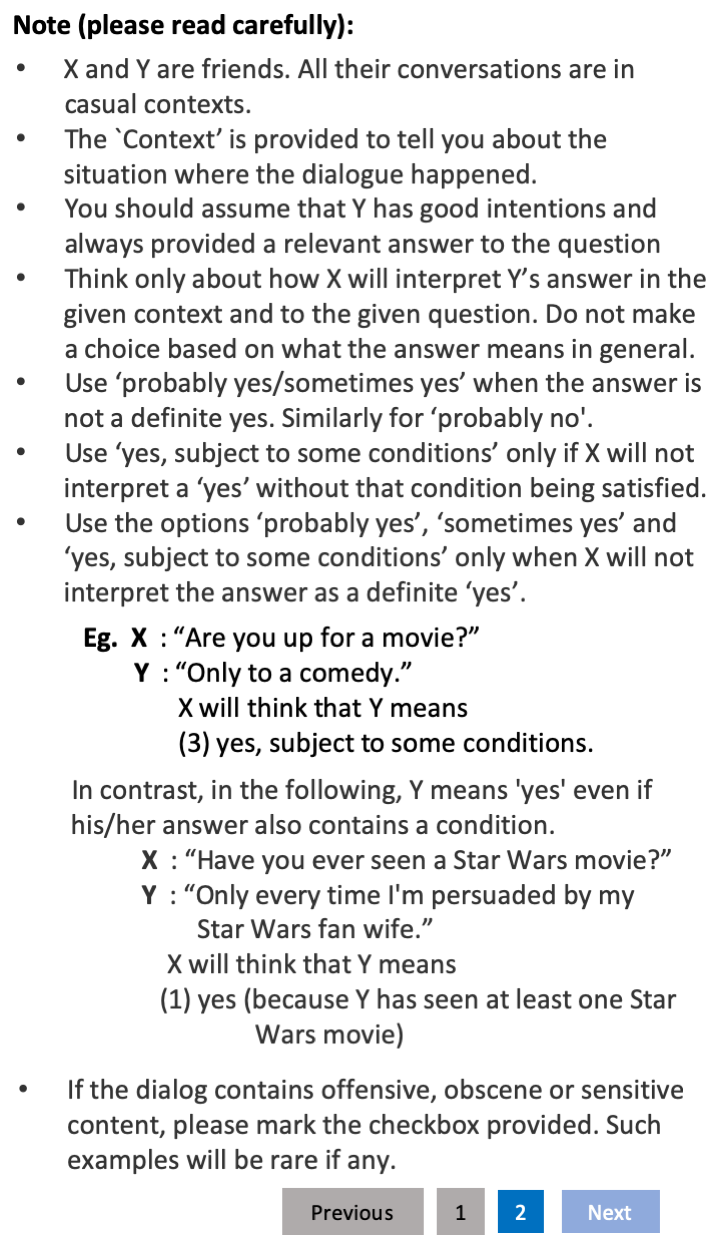}}
    \caption{Additional notes}
  \end{subfigure}\vspace{5mm}
   \begin{subfigure}[b]{0.5\linewidth}
    \fbox{\includegraphics[width=0.9\linewidth]{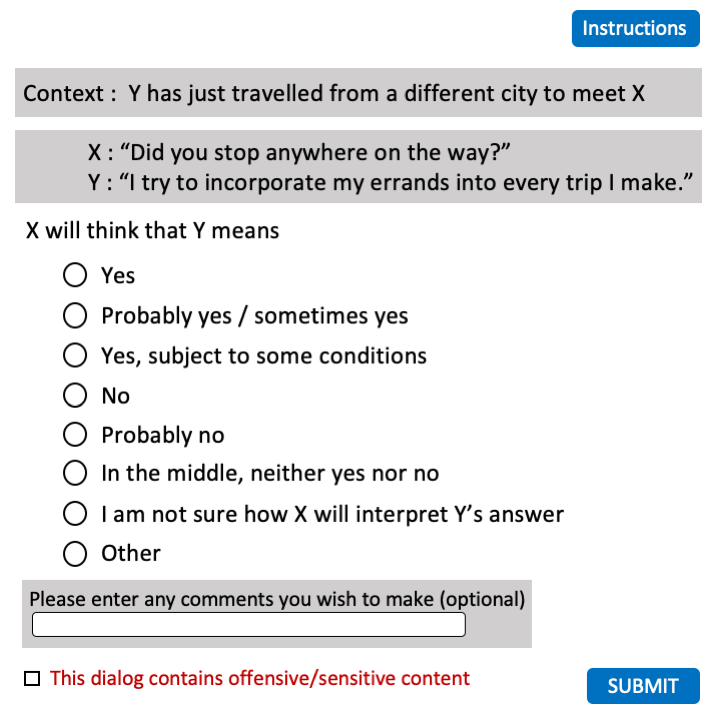}}
    \caption{Example item}
  \end{subfigure}
  \caption{Annotator interface for marking interpretation.}
  \label{fig:step3}
\end{figure*}

\paragraph*{\bf Marking interpretations.} Finally, the question-answer pairs are annotated with meaning categories. Our complete
instructions and annotator interface are in Figure \ref{fig:step3}. Again, the scenario descriptions are modified for the task,
and are given in Table \ref{tab:scenarios_step3}.

This step is fairly complex, so every annotator also worked through 8 practice questions before starting the annotation. After they answered them, the correct answers were shown along with an explanation. These examples are given in Table \ref{tab:practice_step3}.

\begin{table*}[h]
\def\arraystretch{1.5}
\centering
\begin{footnotesize}
\begin{tabular}{p{7cm} | p{7cm}}
(1) Suppose that you are trying to learn about a friend's food preferences, so that you can recommend a local restaurant, but can only ask yes/no questions. Provide 5 useful questions that can be answered "Yes" or "No". & 
(5) On a Friday evening, you are leaving work and see your friend (and colleague) also at the door ready to leave. Provide 5 questions you might ask him/her that can be answered "Yes" or "No".\\
(2) Suppose that you are trying to learn what sorts of activities a friend likes to do during the weekend, so that you can recommend local activities that might interest them, but can only ask yes/no questions. Provide 5 useful questions that can be answered "Yes" or "No". & 
(6) Suppose that you are trying to learn about a friend’s interests related to music. For instance, you could ask about music tastes, instruments played, events they go to, etc. You are only allowed to ask yes/no questions. Provide 5 questions you might ask.\\
(3) Suppose that you are trying to learn what sorts of books someone likes to read, but are only allowed to ask yes/no questions. Provide 5 questions you might ask about books that can be answered "Yes" or "No". &
(7) Your friend has arrived from out of town to visit you. Provide 5 questions you might ask your friend when he/she arrives and during your time together. You are only allowed yes/no questions.\\
(4) Suppose that you are meeting your new neighbour for the first time. You want to find out more about him/her, but you are only allowed to ask yes/no questions. Provide 5 questions you might ask him/her that can be answered "Yes" or "No". &
(8) Suppose that you are at a cafe, and you run into your childhood neighbour. You haven’t seen each other or had any contact for many years. Ask a few questions of your childhood neighbour. You are only allowed yes/no questions. Provide 5 questions you might ask. \\
(9) Suppose that your friend tells you he is thinking of buying a flat  in New York. In this context, provide 5 questions you might ask him/her. You are only allowed to ask yes/no questions.  &
(10) Your friend (not a colleague) is considering switching his/her job. You do not know much about the aspects of his/her job. Ask a few yes/no questions in this context. Provide 5 yes/no questions. \\
\end{tabular}
\end{footnotesize}
\caption{Descriptions of the 10 scenarios in the question collection step.}
\label{tab:scenarios_step1}
\end{table*}

\begin{table*}
\def\arraystretch{1.5}
\centering
\begin{footnotesize}
\begin{tabular}{p{7cm} | p{7cm}}
(1) Imagine that you are talking to a friend. Your friend asks you a question to know more about your food preferences. Answer your friend’s question without using the words ‘yes’ or ‘no’. & 
(5) Imagine that at the end of the week you are leaving work and see your friend (and colleague) also at the door ready to leave. Your friend asks you a question. Answer your friend’s question without using the words ‘yes’ or ‘no’. \\
(2) Imagine that you are talking to a friend. Your friend asks you a question to know more about what activities you like to do during weekends. Answer your friend’s question without using the words ‘yes’ or ‘no’. &
(6) Imagine you are talking to a friend. Your friend asks you a question to know more about your interests related to music. Answer your friend’s question without using the words ‘yes’ or ‘no’.\\
(3) Imagine that you are talking to a friend. Your friend asks you a question to know more about what sorts of books you like to read. Answer your friend’s question without using the words ‘yes’ or ‘no’. &
(7) Imagine that you have just travelled from a different city to visit your friend. Upon your arrival, your friend asks you a question. Answer your friend’s question without using the words ‘yes’ or ‘no’.\\
(4) Imagine that you have just moved into a neighbourhood. One of your new neighbours is a friendly person, and asks you a question. Answer your neighbour’s question without using the words ‘yes’ or ‘no’. &
(8) Imagine that you run into your childhood neighbour at a cafe. You haven’t seen each other or had any contact for many years. Your childhood neighbour asks you a question. Answer your neighbour’s question without using the words ‘yes’ or ‘no’.\\
(9) Imagine that you have just told your friend that you are thinking of buying a flat in New York. Your friend asks you a question to know more about your plans. Answer your friend’s question without using the words ‘yes’ or ‘no’.&
(10 Imagine that you have just told your friend that you are considering switching your job. Your friend asks you a question to know more about your plans. Answer your friend’s question without using the words ‘yes’ or ‘no’.\\
\end{tabular}
\end{footnotesize}
\caption{Descriptions of the 10 scenarios as used in the answer elicitation step.}
\label{tab:scenarios_step2}
\end{table*}

\clearpage

\begin{table*}[!]
\def\arraystretch{1.5}
\centering
\begin{footnotesize}
\begin{tabular}{p{7cm} | p{7cm}}
(1) X wants to know about Y's food preferences. &
(5) X and Y are colleagues who are leaving work on a Friday at the same time. \\
(2) X wants to know what activities Y likes to do during weekends. &
(6)	X wants to know about Y's music preferences. \\
(3) X wants to know what sorts of books Y likes to read. &
(7) Y has just travelled from a different city to meet X. \\ 
(4) Y has just moved into a neighbourhood and meets his/her new neighbour X.  &
(8) X and Y are childhood neighbours who unexpectedly run into each other at a cafe. \\
(9) Y has just told X that he/she is thinking of buying a flat in New York. &
(10) Y has just told X that he/she is considering switching his/her job. \\
\end{tabular}
\end{footnotesize}
\caption{Descriptions of the 10 scenarios as used in the interpretation marking step.}
\label{tab:scenarios_step3}
\end{table*}

\begin{table*}[!]
\centering
\begin{footnotesize}
\begin{tabular}{p{12cm}}
(1) Context: X wants to know about Y's food preferences.\\	
X: ``Do you eat red meat?''	\\
Y: ``I am a vegetarian.''	\\
Answer: No (Vegetarians do not eat meat. So X will interpret it as a ‘no’ answer.) \\ 
\\\hline
(2) Context: X wants to know about Y's food preferences.\\
X: ``Have you had bulgogi?''\\
Y: ``I am not sure.''	\\
Answer: In the middle, neither yes nor no. (Here Y’s response is non-committal. So X’s best option is to interpret it as neither a ‘yes’ nor a ‘no’)\\ 
\\\hline
(3) Context: Y has just travelled from a different city to meet X.	\\
X: ``Did you stop anywhere on the way?''\\
Y: ``I had to get gas a few times.''	\\
Answer: Yes	(X will infer a definite ‘yes’ because Y stopped for gas.)	\\
\\\hline
(4) Context: Y has just travelled from a different city to meet X.	\\
X: ``Did you stop anywhere on the way?''	\\
Y: ``I try to incorporate my errands into every trip I make.''	\\
Answer: Probably yes / sometimes yes (In response to the question, Y mentions that his general tendency is to do errands on the way. So it is likely that he did the same on this trip. Hence ‘probably yes’)\\	
\\ \hline
(5) Context: X wants to know about Y's music preferences.	\\
X: ``Would you go to a punk rock show?''	\\
Y: ``It depends on who is playing.''	\\
Answer: Yes, subject to some conditions (Y might go depending on the artist who is performing. Hence X will interpret that Y is posing a condition for going to the punk show.)	\\ 
\\ \hline		
(6) Context: X wants to know about Y's music preferences.	\\
X: ``Would you go to a punk rock show?''	\\
Y: ``I think I'd enjoy something less chaotic a little more.''	\\
Answer: Probably no (Y’s indicates that he’d prefer some other activity. But he does not completely rule out the possibility of going to a punk show. Hence X will interpret it as ‘probably no’.)\\ 
\\ \hline		
(7) Context: X and Y are colleagues who are leaving work on a Friday at the same time.	\\
X: ``Is you department busy this time of year?''	\\
Y: ``We usually end up working overtime.''\\
Answer: Yes	(Since Y is regularly working overtime, X will infer that Y’s department is busy.)	\\ 
\\ \hline		
(8) Context: X and Y are colleagues who are leaving work on a Friday at the same time.	\\
X: ``Is your department busy this time of year?''\\
Y: ``Just as busy as the rest of the year.''	\\
Answer: In the middle, neither yes nor no (Since it is not known how busy Y’s department is in general, it is unclear if Y is busy at this time. In such cases pick ‘in the middle, neither yes nor no’. Note that if X had background knowledge that Y’s department is usually never busy, he could interpret the answer as a ‘no’ or vice versa. You do not have to assume such information is available when providing your answer.)\\ \hline	
\end{tabular}
\end{footnotesize}
\caption{The 8 practice questions used to train annotators for marking interpretations.}
\label{tab:practice_step3}
\end{table*}

\end{document}